\documentclass[10pt,twocolumn,letterpaper]{article}

\usepackage{cvpr}
\usepackage{times}
\usepackage{epsfig}
\usepackage{graphicx, subfigure}
\usepackage{amsmath, bm}
\usepackage{amssymb}
\usepackage{footnote}
\usepackage{authblk}
\usepackage{amsthm}

\usepackage{xcolor,colortbl}
\definecolor{Gray}{gray}{0.85}
\definecolor{LightCyan}{rgb}{0.88,1,1}
\newcolumntype{m}{>{\columncolor{Gray}}c}
\newcolumntype{p}{>{\columncolor{green!30}}c}
\newcolumntype{r}{>{\columncolor{blue!30}}c}
\newcolumntype{f}{>{\columncolor{red!30}}c}

\usepackage[breaklinks=true,bookmarks=false]{hyperref}

\cvprfinalcopy 

\def\cvprPaperID{****} 

\title{NM-Net: Mining Reliable Neighbors for Robust Feature Correspondences}
\author[1]{Chen Zhao}
\author[1]{Zhiguo Cao}
\author[1]{Chi Li}
\author[2]{Xin Li}
\author[1]{Jiaqi Yang \thanks{Corresponding author}}
\affil[1]{School of Artificial Intelligence and Automation, Huazhong University of Science and Technology}
\affil[2]{Lane Department of Computer Science and Electrical Engineering, West Virginia University}
\affil[1]{\tt\small $\left\lbrace \emph{hust\_zhao, zgcao, li\_chi, jqyang} \right\rbrace$ @hust.edu.cn}
\affil[2]{\tt\small Xin.Li @mail.wvu.edu}

\begin{document}

\maketitle

\begin{abstract}
   Feature correspondence selection is pivotal to many feature-matching based tasks in computer vision. Searching for spatially $k$-nearest neighbors is a common strategy for extracting local information in many previous works. However, there is no guarantee that the spatially $k$-nearest neighbors of correspondences are consistent because the spatial distribution of false correspondences is often irregular. To address this issue, we present a \emph{compatibility}-specific mining method to search for consistent neighbors. Moreover, in order to extract and aggregate more reliable features from neighbors, we propose a hierarchical network named \emph{NM-Net} with a series of convolution layers taking the generated graph as input, which is insensitive to the order of correspondences. Our experimental results have shown the proposed method achieves the state-of-the-art performance on four datasets with various inlier ratios and varying numbers of feature consistencies.
\end{abstract}

\section{Introduction}
\begin{figure}[t]
	
	\centering
	\subfigure[Image segmentation]
	{ \includegraphics[width=0.47\linewidth]{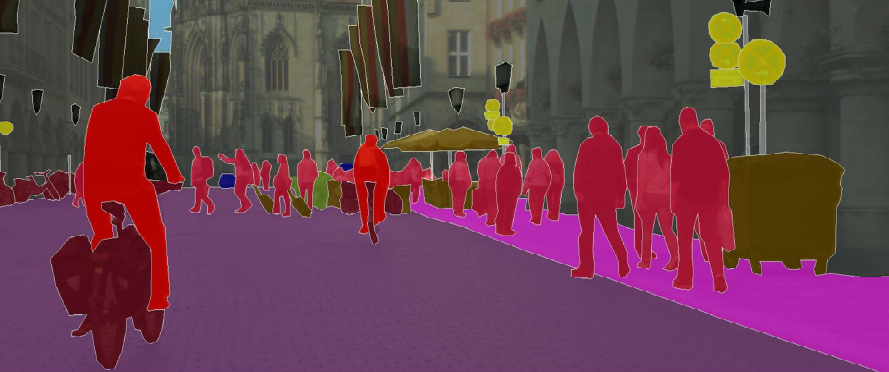}\label{fig_1a} }
	\subfigure[Point cloud segmentation]
	{ \includegraphics[width=0.47\linewidth]{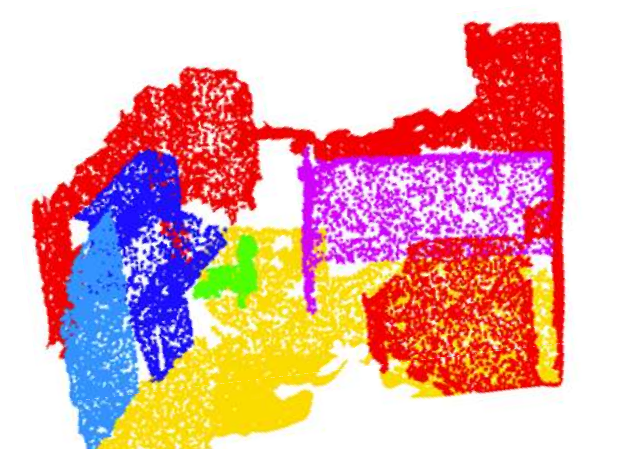}\label{fig_1b} }	
	\subfigure[Correspondence selection]
	{ \includegraphics[width=1.0\linewidth]{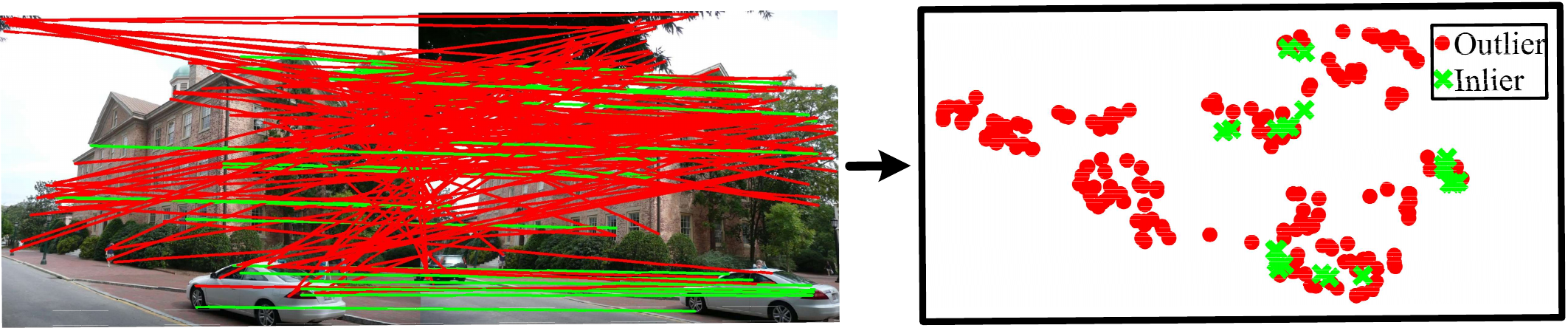}\label{fig_1c} }
	
	\caption{Comparison of (c) correspondence selection (can be viewed as a binary classification task for each correspondence) with standard segmentation problems including (a) image segmentation and (b) point cloud segmentation. Most spatially adjacent pixels and points in (a) and (b) are semantically consistent (belonging to the same class), yet the spatial distribution of mismatches in (c) is irregular, resulting in many outliers (red dots) contaminated in the local regions around inliers (green crosses). For visualization, image correspondences (4D) are projected to a 2D space via $tsne$~\cite{maaten2008visualizing}. }
	\label{fig:1}
	\label{fig:onecol}
\end{figure}

Searching for good feature correspondences (a.k.a. matches) is a fundamental step in computer vision tasks - \eg, structure from motion~\cite{snavely2008modeling}, simultaneous location and mapping~\cite{benhimane2004real}, panoramic stitching~\cite{brown2007automatic}, and stereo matching ~\cite{hirschmuller2008stereo}. Finding consistent feature correspondences between two images relies on two key steps~\cite{lowe2004distinctive, bian2017gms, Ma2017Locality} - \ie, \emph{feature matching} and \emph{correspondence selection}. Specifically, initial correspondences can be obtained by matching local key-point features such as SIFT~\cite{lowe2004distinctive}. Due to various reasons (\eg, key-point localization errors, limited distinctiveness of local descriptors, and illumination/viewpoint changes), mismatches are often inevitable. To address this issue, correspondence selection can be employed as a postprocessing step to ensure correct matches and improve the accuracy~\cite{bian2017gms}. This paper focuses on a \emph{learning}-based approach toward selecting correct matches from an initial set of feature correspondences~\cite{yi2018learning}.

Feature correspondence selection is challenging due to the scarcity of available information as well as the limitation of local feature representations. Spatial positions of matched features are discrete and irregular (note that RGB or texture information is not available any more). To effectively mine the consistency from raw positions, spatially local information is often employed in previous hand-crafted algorithms~\cite{Liu2010Common, bian2017gms, Ma2017Locality}. Indeed, spatially local information has played an important role in image segmentation~\cite{long2015fully} and point cloud segmentation~\cite{qi2017pointnet++}. As shown in Fig.~\ref{fig:1} (a) and (b), most feature points located in adjacent regions are semantically consistent (belonging to the same class). However, spatially local information is unreliable for correspondence selection due to irregular distribution of mismatches. As shown in Fig.~\ref{fig:1} (c), around the vicinity of correct correspondences (marked by green crosses), a large number of mismatches (denoted by red dots) can be found. To overcome this difficulty, we present a compatibility-specific neighbor mining algorithm to search for top-$k$ consistent neighbors of each correspondence. Correspondences are determined to be compatible if they meet the same underlying constraint~\cite{fischler1981random, Albarelli2010Robust, Chen2015Co}. When compared with spatially $k$-nearest neighbor ($knn$) search, the proposed neighbor mining approach is more reliable because potential inliers exhibit guaranteed consistency with each other~\cite{Leordeanu2005A}.

Besides neighbor mining, another important issue is to find a proper representation for correspondence selection. Representations learned by various convolution neural networks (CNN) have become the standard in many computer vision tasks~\cite{Krizhevsky2012ImageNet, simonyan2014very, he2016deep}. Correspondence selection can also be regarded as a binary classification problem for each match - \ie, correct (inlier) vs. false (outlier). Nevertheless, it is often impractical to directly use a CNN to extract features from unordered and irregular correspondences. The first learning-based method for correspondence selection using multi-layer perceptron is proposed recently in~\cite{yi2018learning}, but unfortunately it has ignored useful local information such as those obtained by \emph{compatibility}-specific neighbor mining (shown to be advantageous in our work). To fill this gap, we propose a hierarchical deep learning network called NM-Net (neighbors mining network), where features are successively extracted and aggregated. Compatibility-specific local information is utilized from two aspects: 1) a graph is generated for each correspondence where the nodes denote compatible neighbors found by our neighbor mining approach; 2) features are extracted and aggregated by a set of convolution layers that take the generated graph as input. 

In a nutshell, the contributions of this paper are as follows:
\begin{itemize}
	\item We suggest that compatibility-specific neighbors are more reliable (have stronger local consistency) for feature correspondences than spatial neighbors.
	\item We propose a deep classification network called NM-Net\footnote{The code will be available at \url{https://github.com/sailor-z/NM-Net}} that fully mines compatibility-specific locality for correspondence selection, with hierarchically extracted and aggregated local correspondences. Our network is also insensitive to the order of correspondences. 
	\item Our method achieves the state-of-the-art performance on comprehensive evaluation benchmarks including correspondences with various proportions of inliers and varying numbers of feature consistencies.
\end{itemize}

\section{Related Work}

\noindent\textbf{Parametric methods.}   Generation-verification is arguably the most popular formulation of parametric methods such as RANSAC~\cite{fischler1981random} and its variations (\eg, PROSAC~\cite{Chum2005Matching}, LO-RANSAC~\cite{Chum2003locally}, and USAC~\cite{Raguram2013USAC}). The consistency (compatibility) of correspondences is searched under a global constraint. Specifically, the generation and verification procedures are alternatively used to estimate a global transformation - \eg, homography matrix or essential matrix. Correspondences consistent with the transformation are selected as inliers. Parametric methods have two fundamental weaknesses: 1) the accuracy of estimated global transformation severely degrades when initial inlier ratio is low~\cite{Li2010Rejecting} because the sampled correspondences may include no inlier; 2) the assumed global transformation is unsuitable for the case of multi-consistency matching~\cite{zhao2018scalable} and non-rigid matching~\cite{Ma2017Locality}.

\noindent\textbf{Non-parametric methods.}  Leveraging local information for correspondence selection is a popular strategy in non-parametric methods. For example, a locality preserving matching algorithm is presented in~\cite{Ma2017Locality}, which assumes that local geometric structures in the vicinity of inliers are invariant under rigid/non-rigid transformations, where the spatially $knn$ search is utilized to represent variations of local structures. The spatially local information is exploited in a statistical manner in \cite{bian2017gms}. The similarity of local regions between two images is measured by the number of correspondences; all correspondences located in the regions are considered inliers if the number is larger than a predefined threshold. Additionally, local compatible information has been explored by other non-parametric approaches - \eg, ~\cite{Albarelli2010Robust} measured the compatibility of each two correspondences as a payoff in a game-theoretic framework, where the probability of correct correspondences is iteratively calculated by ESS's algorithm~\cite{weibull1997evolutionary}; ~\cite{Leordeanu2005A} estimated an affinity matrix to represent the compatibility of correspondences and proposed a spectral technique to select inliers. Although these algorithms involve the compatibility information among correspondences, they do not sufficiently mine local information from compatible correspondences. By contrast, we use compatibility-specific neighbors to integrate local information to each correspondence via a data-driven manner.

\noindent\textbf{Learning-based methods.} Deep learning has achieved great success in recent years - \eg, image classification~\cite{Krizhevsky2012ImageNet, simonyan2014very, he2016deep}, object detection~\cite{Ross2014Rich, girshick2015fast, ren2015faster}, and image segmentation~\cite{long2015fully}. However, directly employing a standard CNN is infeasible to  correspondence selection because correspondence representations are irregular and unordered. In ~\cite{yi2018learning}, a deep learning framework based on multi-layer perceptron is employed to find inliers but without any local feature extraction or aggregation. Considering that point-cloud data has the similar characteristics with correspondences, PointNet~\cite{qi2017pointnet} and PointNet++~\cite{qi2017pointnet++} can be referred for correspondence selection, which have been developed for point cloud classification and segmentation recently. Nonetheless, each point is individually processed in PointNet without any local information involved;  spatially nearest information is exploited by PointNet++ in a grouping layer even though such spatially local information could be unreliable for correspondences. Different from these learning-based methods for irregular data, our approach covers both locality selection and locality integration concerns via a compatibility metric and a hierarchical manner, respectively.

\section{Motivation}
\label{sec:moti}
Finding consistency (compatibility) among matches and selecting good correspondences is a chicken-and-egg problem:  finding the consistency (for an assumed global transformation as an example) would require a set of inliers (\ie, the knowledge about good correspondences); but meantime, the selection of inliers also relies on the results of finding reliable consistency. To get around this circular problem, we propose to utilize local information of correspondences as an surrogate representation for feature consistency.

Local information has been the cornerstone in many learning-based methods for image/point cloud classification and segmentation\cite{simonyan2014very, he2016deep, long2015fully, qi2017pointnet}, where local context features in convolution kernels are commonly extracted. Since correspondence selection can be considered as a binary classification problem for each correspondence (\ie, inlier vs. outlier), it appears plausible to mine reliable local information for establishing good correspondences.
\begin{figure}[t]
	\centering
	\subfigure[Spatially $k$-nearest neighbors]
	{ \includegraphics[width=0.9\linewidth]{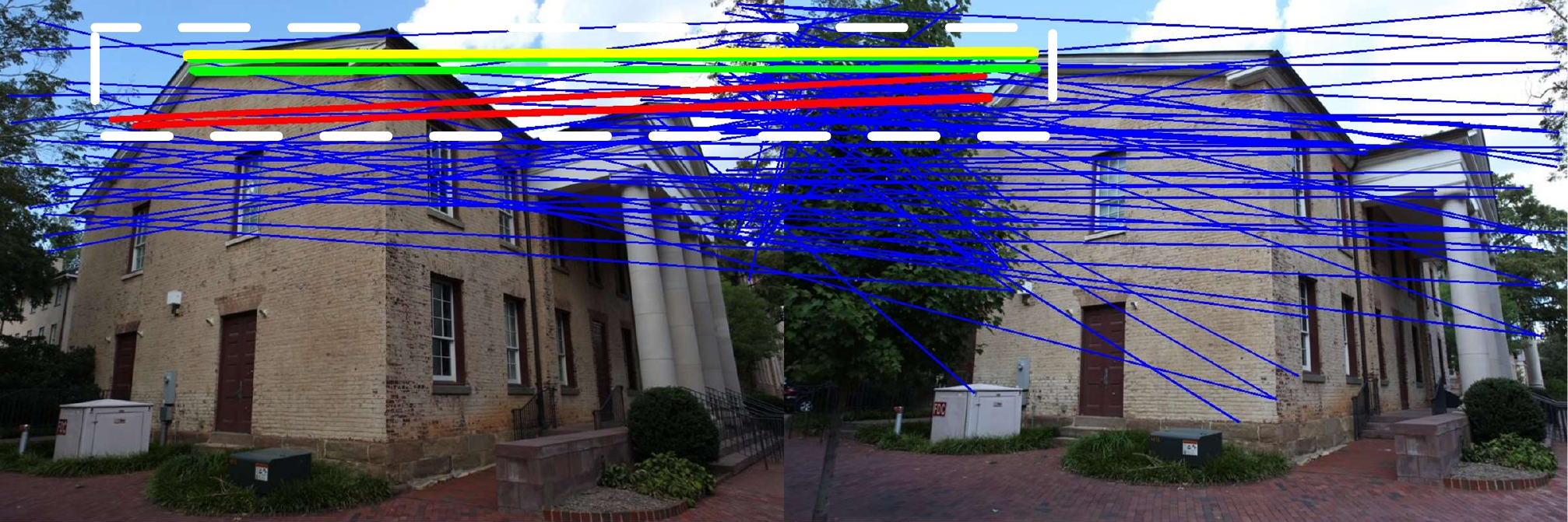}\label{fig_3b} }
	\subfigure[Compatibility-specific $k$-nearest neighbors]
	{ \includegraphics[width=0.9\linewidth]{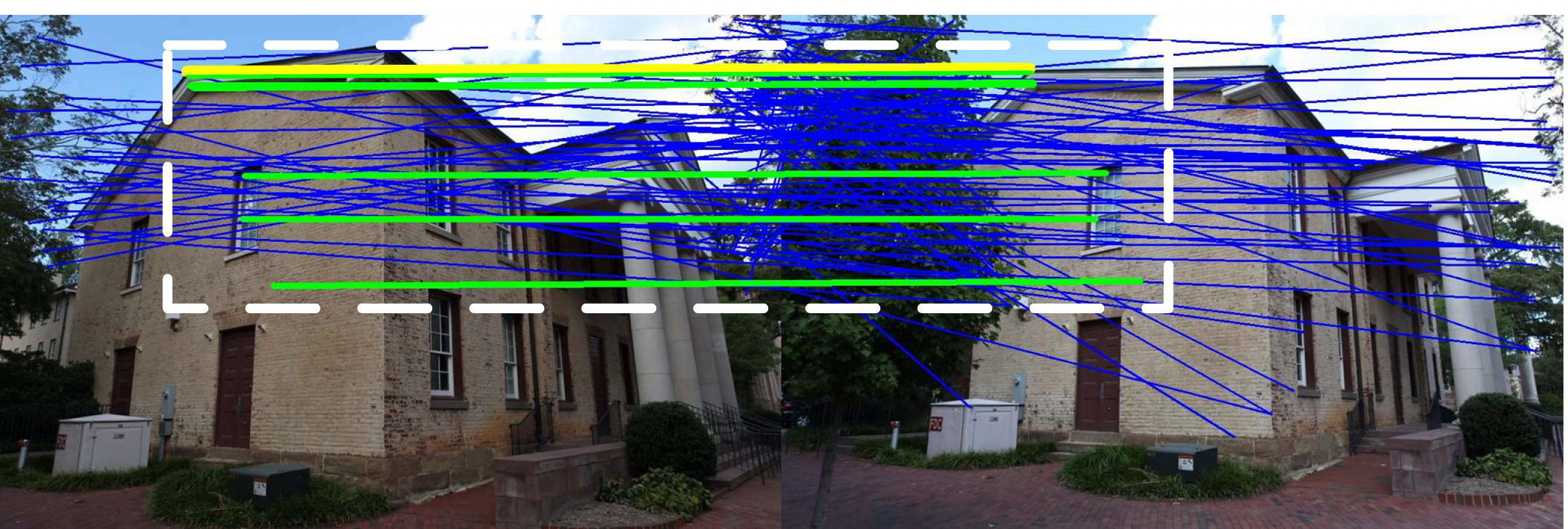}\label{fig_3c} }	
	\caption{Visual illustration of (a) spatially $k$-nearest neighbors and (b) compatibility-specific $k$-nearest neighbors of feature correspondences. The blue lines represent initial feature correspondences between two images, the yellow line denotes a sampled inlier, and the green lines and red lines respectively imply the neighbors of the sampled inlier being inliers and outliers. Two outliers are included in (a) the spatially k-nearest neighbors, which are considered as inconsistent portions; (b) the compatibility-specific $k$-nearest neighbors are consistent without outliers, but their positions are not necessarily to be spatially close.}
	\label{fig:2}
	\label{fig:onecol}
\end{figure}
\begin{figure}[t]
	\centering
	\subfigure[$< 20\%$]
	{ \includegraphics[width=0.47\linewidth]{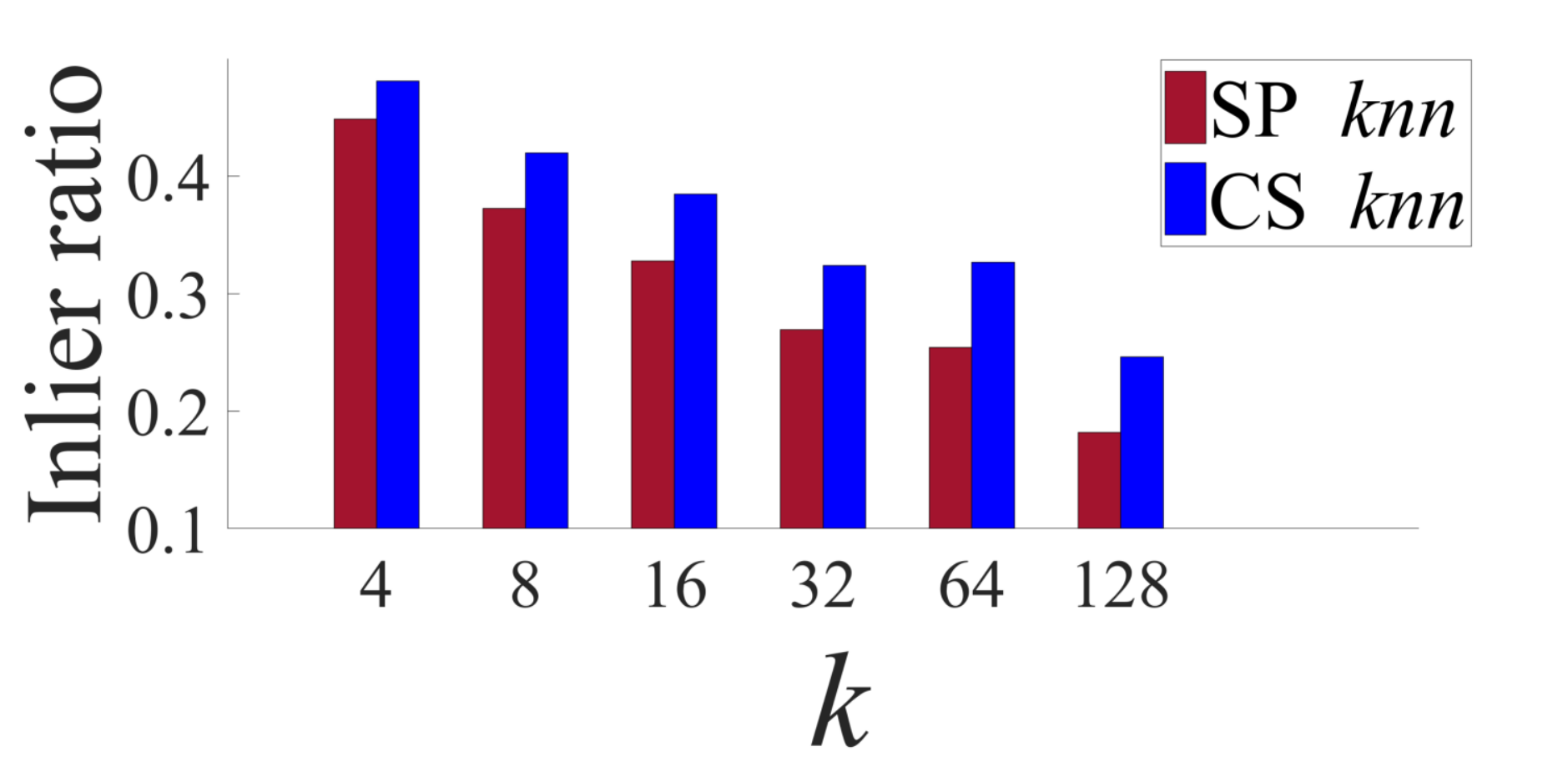}\label{fig_2a} }
	\subfigure[$20 - 35\%$]
	{ \includegraphics[width=0.47\linewidth]{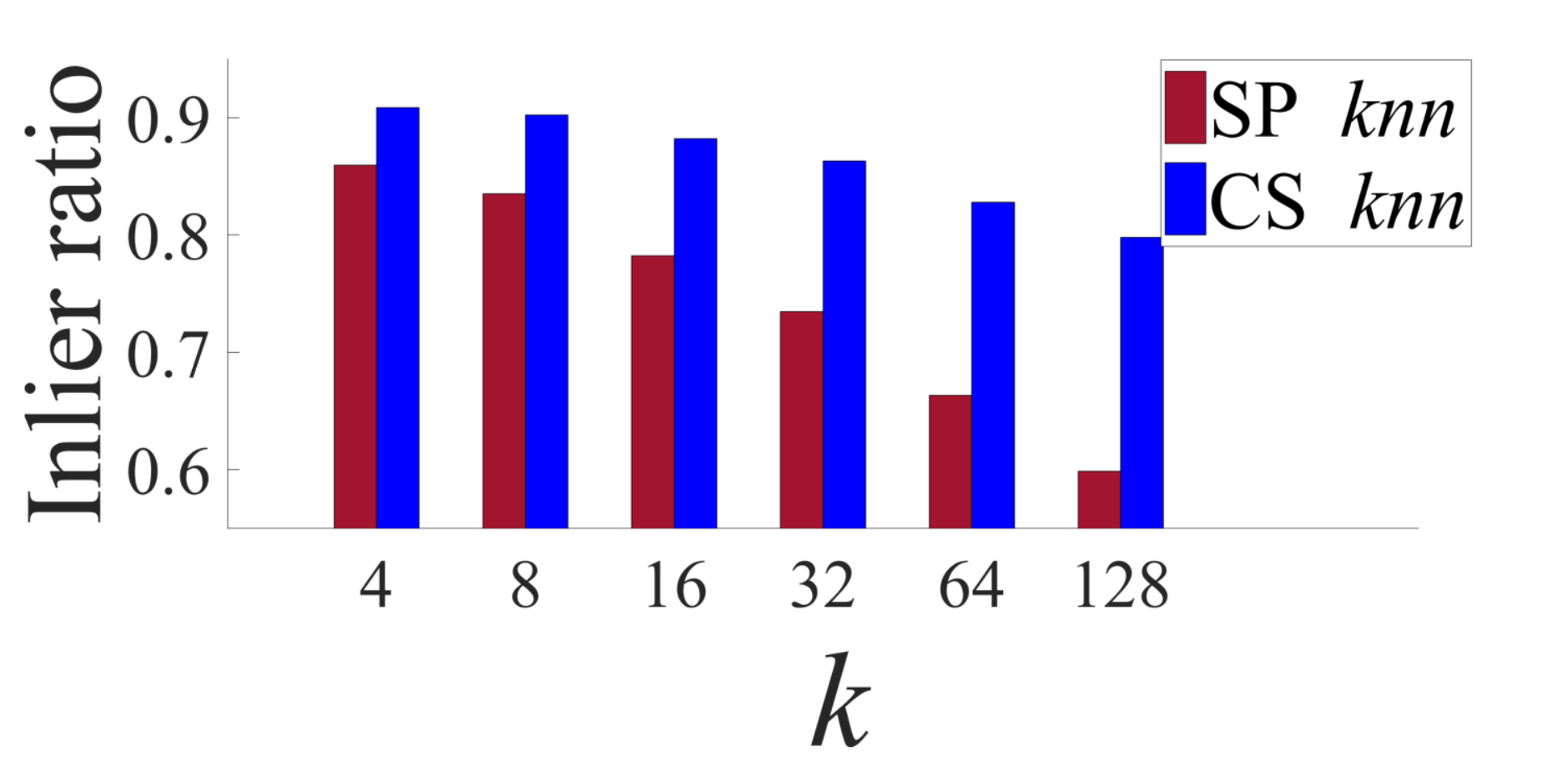}\label{fig_2b} }	
	\subfigure[$35 - 50\%$]
	{ \includegraphics[width=0.47\linewidth]{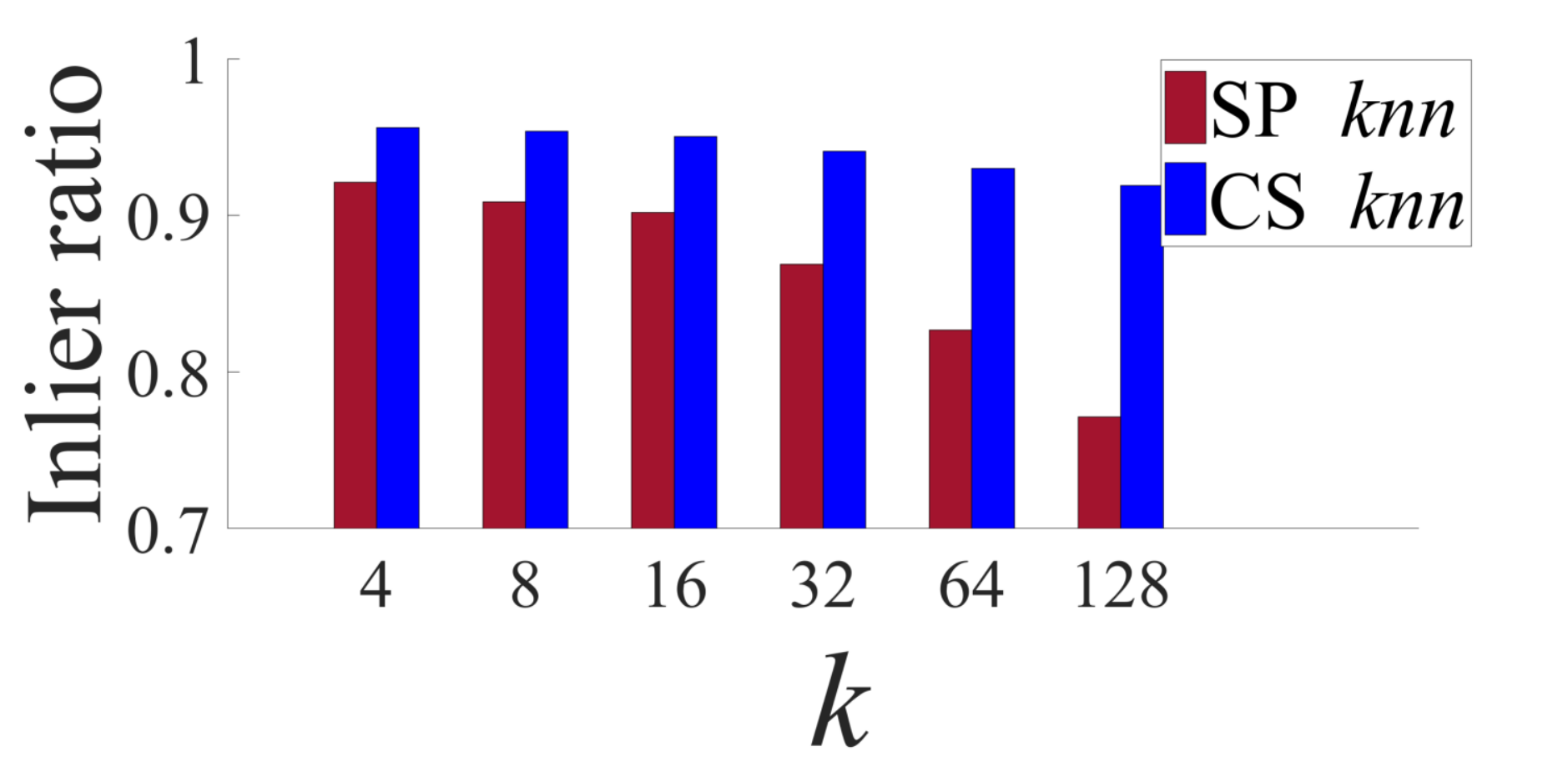}\label{fig_2c} }
	\subfigure[$> 50\%$]
	{ \includegraphics[width=0.47\linewidth]{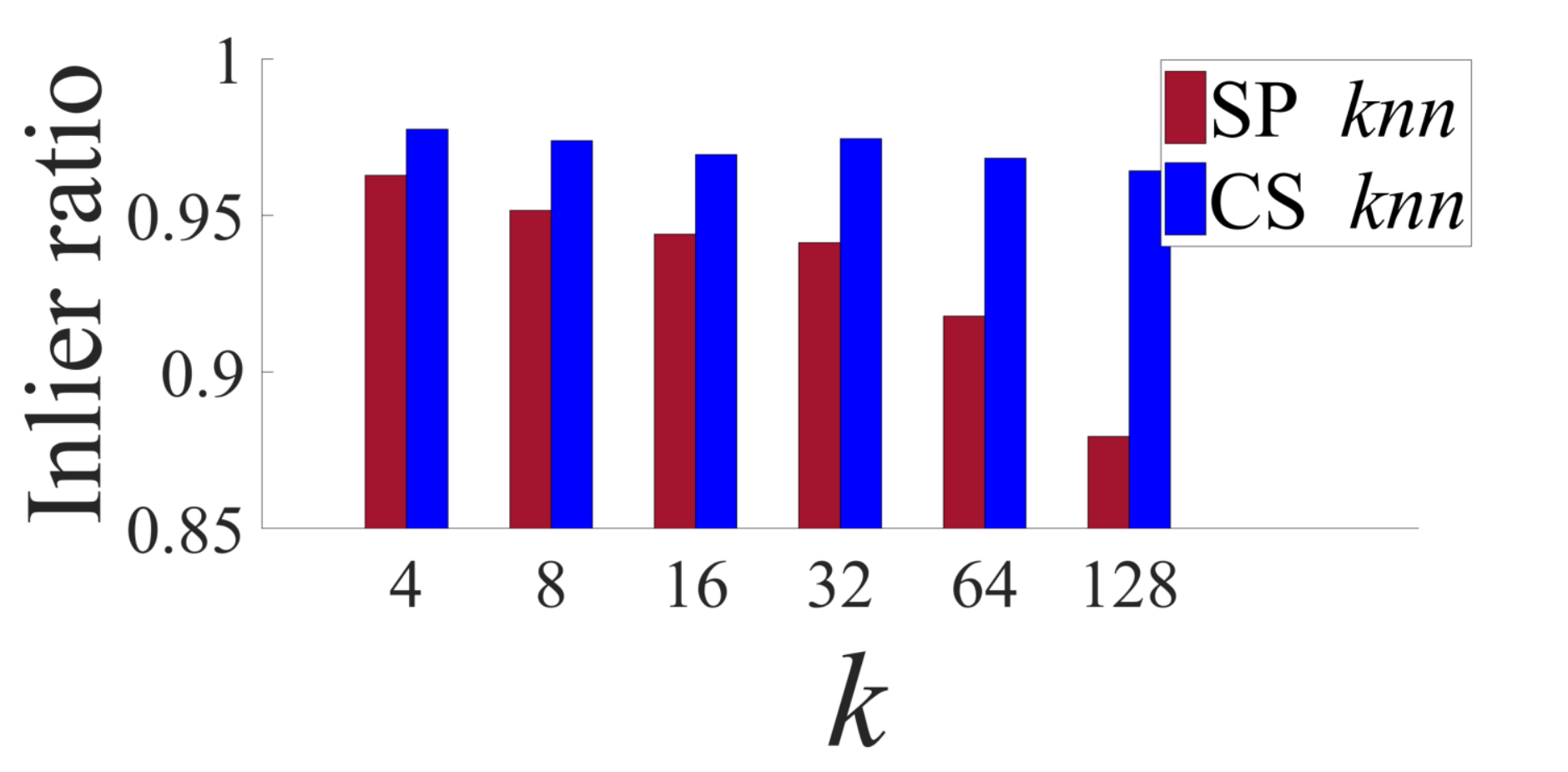}\label{fig_2d} }	
	\caption{Comparison between the spatially $knn$ (SP $knn$) and the compatibility-specific $knn$ (CS $knn$). Hundreds of image pairs are sampled from our experimental datasets (Sect.~\ref{sec:exper}) and divided into $4$ parts according to inlier ratios of the initial correspondence sets, - \ie, (a) lower than 20\%, (b) from 20\% to 35\%, (c) from 35\% to 50\%, and (d) higher than 50\%. The assessment metric is the average inlier ratio of the neighbors of correct correspondences.}
	\label{fig:3}
	\label{fig:onecol}
\end{figure}
As aforementioned, hand-crafted methods have employed spatially local information to select correct matches. However, unlike the cases of images and point clouds, directly using spatially local information for feature correspondence selection is not a good idea as shown in Fig.~\ref{fig:2} (a) - the spatially selected $k$-nearest neighbors of an inlier are incompatible with two outliers. By contrast, the neighbors picked by a compatibility metric (which will be formally defined in the next section) are consistent as shown in Fig.~\ref{fig:2} (b), but their positions are not necessarily to be spatially adjacent to the query correspondence. Since the matched keypoints of an inlier indicate the same 3D position from different viewpoints in the real world, the consistency among inliers is readily guaranteed. Such observation motivates us to develop a compatibility-specific method to mine consistent neighbors.

The superiority of compatibility-specific neighbor mining can be further justified by the statistics collected from experimental datasets (Sect.~\ref{sec:exper}) as shown in Fig.~\ref{fig:3}. The percentages of inliers in compatibility-specific nearest neighbors are remarkably higher than those results in spatially nearest neighbors in all cases, with the gap being more dramatic as $k$ increases. These empiric findings strongly suggest that neighbors chosen by the compatibility-specific method are more reliable.

\section{Method}
\label{sec:met}

In addition to neighbor mining, correspondence selection requires an appropriate representation of the information provided by reliable neighbors. Built upon the success of deep learning in many visual recognition tasks~\cite{he2016deep, ren2015faster, long2015fully}, a learning-based method was developed for correspondence selection in~\cite{yi2018learning}. A key strategy behind~\cite{yi2018learning} is to use multi-layer perceptron that individually processes unordered correspondences. Unfortunately, this recent work fails to integrate local information for each correspondence; a key new insight of our work is to demonstrate the benefits of exploiting locality for correspondence selection by our proposed NM-Net. Our framework employs ResNet~\cite{he2016deep} as the backbone and compatibility-specific neighbor mining algorithm as the grouping module. The 4D raw correspondences are taken as input and the classification of each correspondence (\ie, inlier or outlier) is output.
\subsection{Problem Statement}
Given a pair of images $(I, I^{'})$, two sets of discrete keypoints $(\mathcal{K}=\{{\mathbf{k}}\}, \mathcal{K^{'}}=\{{\mathbf{k}^{'}}\})$ are detected and local patterns around those keypoints are described as $(\mathcal{P}=\{{\mathbf{p}}\}, \mathcal{P^{'}}=\{{\mathbf{p}^{'}}\})$. The initial correspondence set $\mathcal{C} = \{c\}$ is generated by brute-force matching between $(\mathcal{K}, \mathcal{K^{'}})$ based on descriptor similarities. The correspondence selection boils down to a binary classification problem focusing on the classification of each $c$ as an inlier $c_{inlier}\in \mathcal{C}_{inlier}$ or an outlier $c_{outlier}\in \mathcal{C}_{outlier}$.
\subsection{Mining Neighbors}
Compatibility-specific neighbor mining plays a crucial role in our network for the following two reasons. First, it explores the local space of each correspondence and extracts local information by our proposed compatibility metric. Second, it integrates unordered correspondences into a graph in which nodes correspond to the mined neighbors so that convolutions can be performed for further feature extraction and aggregation. As mentioned in Sect.~\ref{sec:moti}, given a pair of correspondences $(c_i, c_j)$, quantifying the compatibility score denoted by $s(c_i, c_j)$ is non-trivial due to lacking of label information. One promising attack is that the variations of local structures around $(\mathbf{k}_i, \mathbf{k}_i^{'})$ and $(\mathbf{k}_j, \mathbf{k}_j^{'})$ are similar if $c_i$ and $c_j$ are compatible~\cite{Albarelli2010Robust}. Based on this important observation, we propose to compute $s(c_i, c_j)$ by exploiting the variations as follows.

First, the Hessian-affine detector~\cite{Mikolajczyk2004Hessian} is used to detect keypoints; it provides the local affine information around keypoints that is required by the introduced compatibility metric. Local affine information is critical for searching for consistent correspondences when images undergo viewpoint or scale changes~\cite{Chen2015Co}. Second, a transformation characterizing the variation between local structures around $(\mathbf{k}_i, \mathbf{k}_i^{'})$ of $c_i$ can be calculated by
\begin{equation}\label{eq:LRF2}
\mathbf{H}_i={\mathbf{T}_{i}^{'}}\mathbf{T}_i^{-1},
\end{equation}
where a pair of $3 \times 3$ matrices $(\mathbf{T}_i, \mathbf{T}_{i}^{'})$ describe the positions and local structures of the matched keypoints. Then we can calculate $\mathbf{T}_i$ by (the same for $\mathbf{T}_{i}^{'}$) 
\begin{equation}\label{eq:LRF3}
\mathbf{T}_i=\left[ \begin{matrix}
\mathbf{A}_i & \mathbf{k}_i  \\
\mathbf{0} & 1  \\
\end{matrix} \right],
\end{equation}
where $\mathbf{A}_i$ is a $2 \times 2$ matrix representing the local affine information extracted by the Hessian-affine detector and $\mathbf{k}_i$ is the position of the keypoint. Third, intuitively, $c_i$ and $c_j$ are compatible if the corresponding transformations $\mathbf{H}_i$ and $\mathbf{H}_j$ are consistent; in other words, local structure variations estimated by a consistent pair of transformations should be similar. Consequently, we have adopted reprojected errors that represent local structure variations to measure the dissimilarity of $(\mathbf{H}_i, \mathbf{H}_j)$ by
\begin{equation}\label{eq:LRF4}
{e_j({c}_i)}=\left|\rho ({\mathbf{H}_{j}}\cdot \left[ \begin{matrix}
{\mathbf{k}_{i}}  \\
1  \\
\end{matrix} \right])-\rho ({\mathbf{H}_{i}}\cdot \left[ \begin{matrix}
{\mathbf{k}_{i}}  \\
1  \\
\end{matrix} \right])\right|,
\end{equation}
where $\rho ({{\left[ \begin{matrix}
		a & b & c  \\
		\end{matrix} \right]}^{T}})={{\left[ \begin{matrix}
		a/c & b/c  \\
		\end{matrix} \right]}^{T}}.
$ Note that the compatibility of $(c_i, c_j)$ is negatively correlated with the sum of reprojected errors $(e_j({c}_i)+e_i({c}_j))$. As a strategy of normalizing $s_{ij}$ to the range of $\left(0, 1 \right] $, we propose to use a Gaussian kernel - \ie,
\begin{equation}\label{eq:LRF6}
s(c_i, c_j)={{e}^{-\lambda (e_j({c}_i)+e_i({c}_j))}},
\end{equation}
where $\lambda$ is a hyper-parameter. Note that $\lambda$ will not affect the ranking of $s(c_i, c_j)$; so the search of compatible neighbors for each correspondence is insensitive to $\lambda$. For any given $c_i$, a graph $\mathcal{G}_i$ is generated by first selecting the neighbors of $c_i$, - \ie, $\left\lbrace c_i^{j}\right\rbrace (j=1,2, ..., k)$, as those correspondences with top-$k$ $s(c_i, c_i^{j})$, and then sequentially linking $c_i$ with all its neighbors.

\subsection{Network Architecture}
\begin{figure}[t]
	\centering
	\subfigure[Network architecture]
	{ \includegraphics[width=0.8\linewidth]{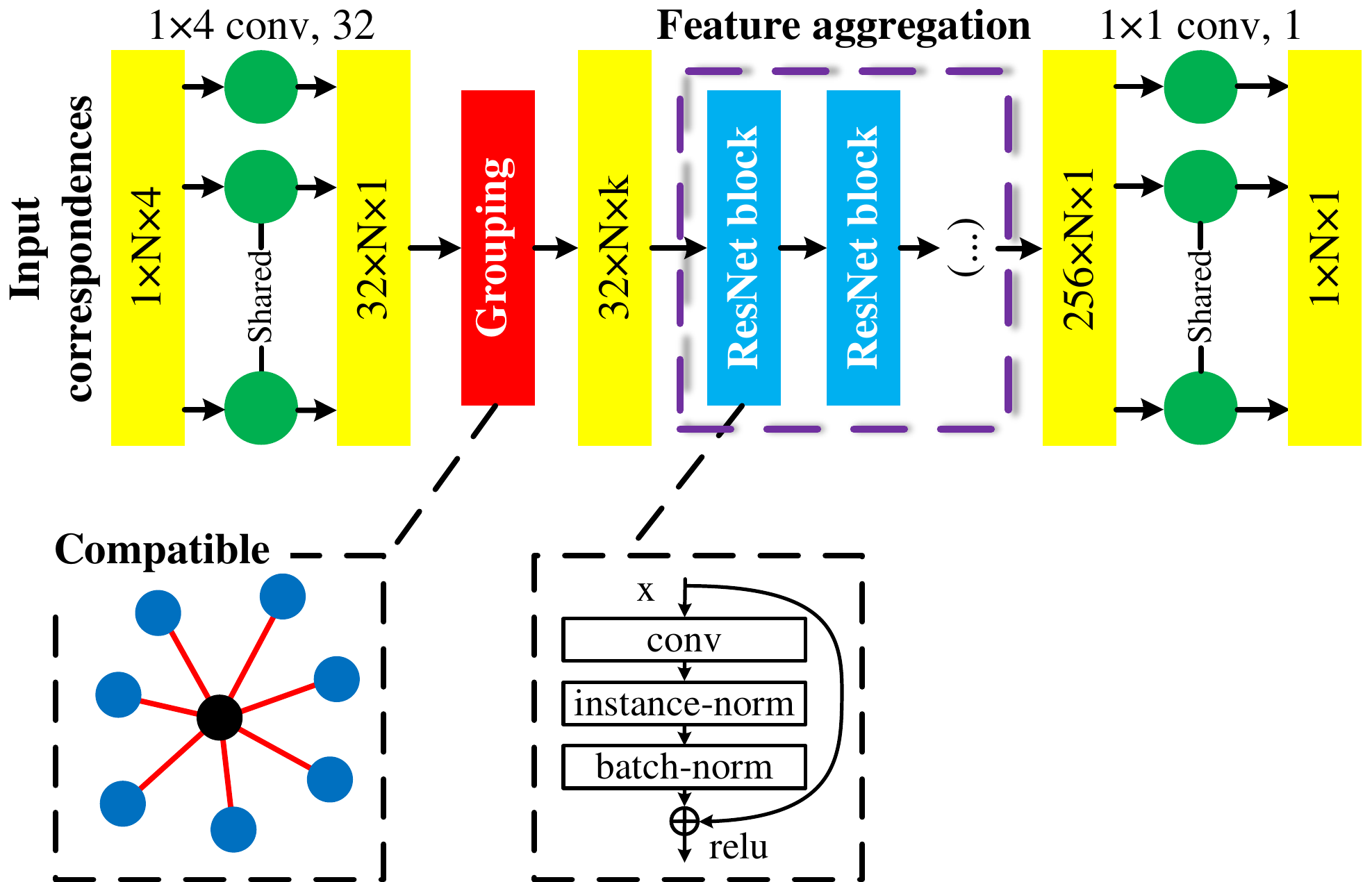}\label{fig_4a} }
	\subfigure[Feature aggregation]
	{ \includegraphics[width=0.8\linewidth]{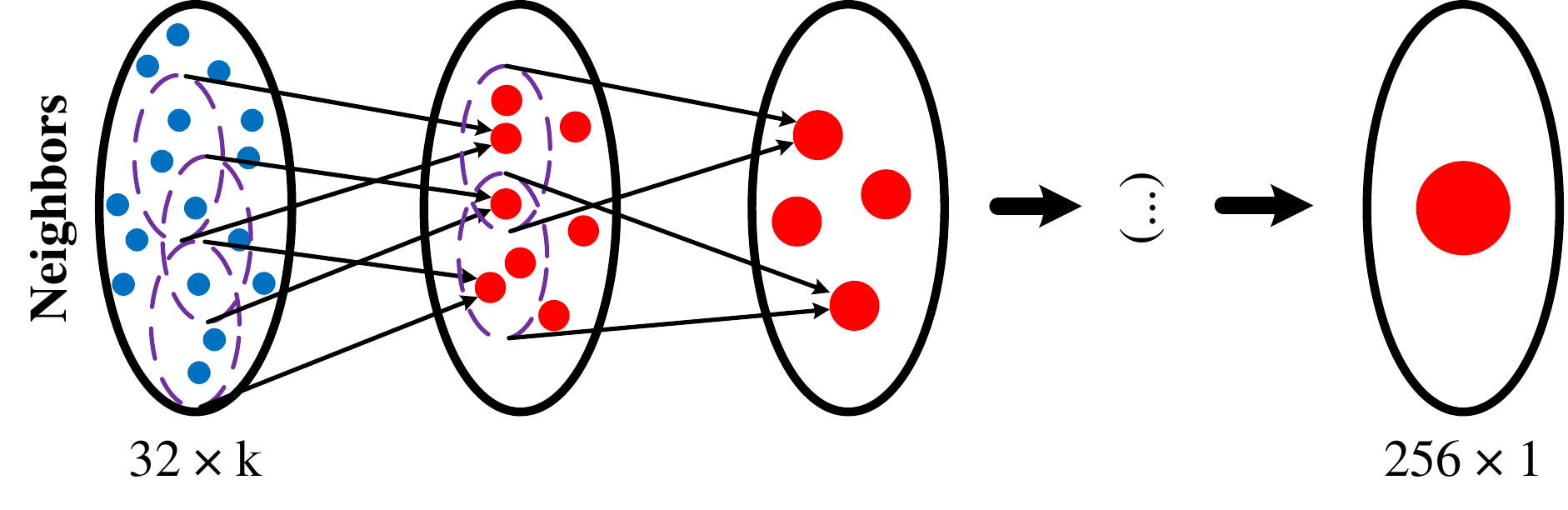}\label{fig_4b} }	
	\caption{{\bf NM-Net architecture.}  NM-Net is (a) a classification network for feature correspondences. A grouping module is designed to first mine reliable neighbors via the compatibility metric and then convert them into a graph for each correspondence to achieve ordered representations. In the bottom-left dashed box of (a), the black dot denotes a query correspondence and those blue dots are its compatibility-specific neighbors. (b) illustrates the hierarchical aggregation of the features extracted from neighbors by a series of ResNet blocks.}
	\label{fig:4}
	\label{fig:onecol}
\end{figure}
The network architecture as shown in Fig.\ref{fig:4} includes two key modules - \ie, \emph{grouping} and \emph{ResNet} blocks. Our network design is partially inspired by hierarchically extracted features to fully leverage correspondence-level local information (\eg, PointNet++~\cite{qi2017pointnet++}). The details of NM-Net are given as follows.

\noindent\textbf{Feature extraction and aggregation.} In NM-Net, features are extracted and aggregated along three lines. First, we use a grouping module to extract local information for each correspondence, where compatibility-specific $knn$ search is adopted. The unordered raw correspondences are converted to $N$ graphs $\mathcal{G}_i(i=1,2, ..., N)$, where the nodes are sorted by $s(c_i, c_i^{j})$, resulting in a set of regular organizations that are invariant to the order of correspondences. Second, features are hierarchically extracted and aggregated by a set of convolutions as
\begin{equation}\label{eq:LRF7}
{\mathbf{f}_j^{n}}=\sum\limits_{j=1}^{d}{{\omega_{j}}\mathbf{f}_{j}^{n-1}},
\end{equation}
where ${\mathbf{f}^{n}_{j}}$ is the feature of the $j$-th node in $\mathcal{G}_i$ at the $n$-th layer, $d$ is the size of convolution kernel, and $\omega_j$ denotes the learned weight. The $N \times k$ feature map is successively aggregated into $N \times 1$, with the feature dimension being increased from $32$ to $256$ as shown in Fig.~\ref{fig:4} (b). In contrast to~\cite{yi2018learning}, our convolutions take regular graphs as input instead of isolated correspondences, with reliably captured local features. Third, for global feature extraction, Instance Normalization~\cite{Ulyanov2017Improved} is used to normalize the feature map in each ResNet block, which has been proven more effective than average-pooling and max-pooling in~\cite{yi2018learning}.

\noindent\textbf{Loss function.} A simple yet effective cross-entropy loss function is used to calculate the deviation between the outputs and corresponding labels - \ie,
\begin{equation}\label{eq:LRF8}
{L(\omega , \mathcal{C})=\frac{1}{N}\sum\limits_{i=1}^{N}{{{\alpha }_{i}}H\left( {{y}_{i}},S\left( g\left( {{c}_{i}},\omega  \right) \right) \right)}},
\end{equation}
where $g$ is the output of NM-Net, $S$ indicates the logistic function, $y_i$ denotes the ground-truth label of $c_i$, $H$ represents a binary cross entropy function, and $\alpha_i$ is a self-adaptive weight to balance positive and negative samples. The regression loss in~\cite{yi2018learning} is not used because the ground-truth global transformation may be unavailable in some applications such as multi-consistency matching and non-rigid matching. As we will show next, our simpler form of loss function can achieve even better performance.
\section{Experiments}
\label{sec:exper}
This section includes extensive experimental evaluations on four standard datasets covering a variety of contexts - \ie, narrow and wide baseline matching, matching for reconstruction (\ie, structure from motion), and matching with multiple consistencies. We also present comprehensive comparisons with several state-of-the-art methods including both hand-crafted approaches (\ie, RANSAC~\cite{fischler1981random}, GTM~\cite{Albarelli2010Robust}, and LPM~\cite{Ma2017Locality}) and learning-based approaches (\ie, PointNet~\cite{qi2017pointnet}, PointNet++~\cite{qi2017pointnet++}, and LGC-Net~\cite{yi2018learning}).
\begin{table*}[t]
	\begin{center}
		\begin{tabular}{|c||c||c||c||c||c||c|}
			\hline
			Dataset & \# Image pairs & \# Training & \# Validation & \# Testing & Inlier ratio (\%) & Challenges\\
			\hline\hline
			NARROW & 24070 & 16849 & 3610 & 3610 & 40.827 & VP changes\\
			WIDE & 11426 & 7998 & 1713 & 1713 & 32.771 & VP changes \\
			COLMAP~\cite{schoenberger2016sfm} & 18850 & 13195 & 2827 & 2827 & 7.496 & VP changes, rotation \\
			MULTI~\cite{zhao2018scalable} & 45 & - & - & - & 40.828 & Dynamic scenarios \\
			\hline
		\end{tabular}
	\end{center}
	\caption{{\bf Properties of the experimental datasets.} VP means viewpoint and the inlier ratio indicates the average proportion of inliers in initial correspondence sets computed over a whole dataset.}
	\label{tab:tab1}
\end{table*}
\begin{table*}[t]
	\centering
	\subtable{
		\begin{tabular}{|c||p||r||f||c||c||c||c||c|}
			\hline
			Method & Precision (\%) & Recall (\%) & F-measure (\%) & MSE & MAE & Median & Max & Min ($10^{-2}$) \\
			\hline\hline
			RANSAC~\cite{fischler1981random} & 86.923 & 60.397 & 69.194 & {\bf2.017} & 2.622 & {\bf2.809} & 4.978 & 0.755\\
			GTM~\cite{Albarelli2010Robust} & 88.707 & 52.949 & 65.653 & 2.042 & 2.728 & 2.886 & 4.968 & 2.467 \\
			LPM~\cite{Ma2017Locality} & 72.667 & 68.504 & 70.173 & 2.087 & 2.879 & 3.107 & 4.869 & 25.453 \\
			PointNet~\cite{qi2017pointnet} & 79.003 & 86.163 & 82.102 & 2.293 & 2.787 & 3.503 & {\bf4.728} & 1.180\\
			PointNet++~\cite{qi2017pointnet++} & 83.677 & 85.045 & 84.112 & 2.248 & 2.773 & 3.328 & 5.128 & 3.899 \\
			LGC-Net~\cite{yi2018learning} & 95.238 & {\bf98.405} & 96.611 & 2.096 & {\bf2.255} & 3.021 & 5.006 & 0.558 \\
			NM-Net-sp & 96.946 & 97.659 & 97.283 & 2.482 & 2.664 & 3.687 & 5.038 & {\bf0.245} \\
			NM-Net & {\bf97.169} & 97.870 & {\bf97.501} & 2.436 & 2.608 & 3.630 & 5.021 & 0.390 \\
			\hline
		\end{tabular}
		\label{tab:tab2a}
	}
	\subtable{
		\begin{tabular}{|c||p||r||f||c||c||c||c||c|}
			\hline
			Method & Precision (\%) & Recall (\%) & F-measure (\%) & MSE & MAE & Median & Max & Min ($10^{-2}$)\\
			\hline\hline
			RANSAC~\cite{fischler1981random} & 80.740 & 51.198 & 60.350 & 2.052 & 2.711 & 3.125 & 5.040 & 1.347\\
			GTM~\cite{Albarelli2010Robust} & 79.989 & 47.711 & 58.881 & 2.040 & 2.784 & 3.068 & 5.041 & 1.046 \\
			LPM~\cite{Ma2017Locality} & 62.940 & 64.487 & 62.828 & {\bf2.038} & 2.921 & 3.080 & 4.926 & 19.782 \\
			PointNet~\cite{qi2017pointnet} & 64.730 & 77.287 & 70.068 & 2.282 & 2.863 & 3.458 & {\bf4.905} & 5.528\\
			PointNet++~\cite{qi2017pointnet++} & 73.926 & 81.856 & 77.245 & 2.180 & 2.771 & 3.255 & 5.013 & 3.020 \\
			LGC-Net~\cite{yi2018learning} & 88.139 & {\bf97.138} & 91.264 & 2.059 & {\bf2.230} & {\bf2.995} & 5.061 & 1.226 \\
			NM-Net-sp & 91.742 & 94.039 & 92.749 & 2.513 & 2.731 & 3.751 & 5.110 & 0.650 \\
			NM-Net & {\bf92.332} & 94.251 & {\bf93.145} & 2.488 & 2.718 & 3.781 & 5.113 & {\bf0.553} \\
			\hline
		\end{tabular}
		\label{tab:tab2b}
	}
	\subtable{
		\begin{tabular}{|c||p||r||f||c||c||c||c||c|}
			\hline
			Method & Precision (\%) & Recall (\%) & F-measure (\%) & MSE & MAE & Median & Max & Min ($10^{-2}$)\\
			\hline\hline
			RANSAC~\cite{fischler1981random} & 25.156 & 14.477 & 17.464 & 1.984 & 2.985 & 3.169 & 5.044 & 1.383\\
			GTM~\cite{Albarelli2010Robust} & 22.931 & 19.913 & 19.075 & 2.004 & 3.073 & 3.245 & 5.102 & 4.804 \\
			LPM~\cite{Ma2017Locality} & 15.879 & 34.293 & 19.595 & 2.019 & 3.037 & 3.113 & {\bf4.693} & 17.298 \\
			PointNet~\cite{qi2017pointnet} & 13.596 & 41.765 & 19.710 & 2.051 & 2.864 & 3.193 & 4.878 & 14.138\\
			PointNet++~\cite{qi2017pointnet++} & 18.659 & 41.953 & 24.301 & 2.060 & 2.902 & 3.200 & 4.877 & 4.150 \\
			LGC-Net~\cite{yi2018learning} & 26.383 & {\bf71.132} & 33.949 & 1.981 & 2.554 & 3.071 & 4.717 & {\bf1.265} \\
			NM-Net-sp & 29.296 & 59.710 & 37.503 & 1.983 & 2.446 & 3.047 & 5.125 & 2.250 \\
			NM-Net & {\bf31.003} & 58.499 & {\bf38.887} & {\bf1.953} & {\bf2.402} & {\bf3.027} & 4.989 & 1.514 \\
			\hline
		\end{tabular}
		\label{tab:tab2c}
	}
	\caption{{\bf Evaluation results.} The three tables from top to bottom are results on NARROW, WIDE, and COLMAP datasets, respectively. Precision, recall, and F-measure are colored for highlighting because they explicitly measure the performance of correspondence selection. NM-Net-sp indicates a variant of our NM-Net with spatial neighbors. The best result in each column is rendered in bold.}
	\label{tab:tab2}
\end{table*}
\subsection{Experimental Setup}

\noindent\textbf{Optimization and architecture details.}  The configuration of NM-Net (Fig.~\ref{fig:4} (a)) is C(32, 1, 4)-GP-R(32, 1, 3)-R(32, 1, 3)-R(64, 1, 3)-R(64, 1, 3)-R(128, 1, 3)-R(128, 1, 3)-R(256, 1, 3)-R(256, 1, 3)-C(256, 1, 1)-C(1, 1, 1), where C($n$, $h, w$) denotes a convolution layer with $n$ output channels and a $h \times w$ convolution kernel, GP indicates the grouping module, and R($n$, $h, w$) represents a ResNet block that includes two convolution layers C($n$, $h, w$). Every convolution layer is followed by Instance Normalization, Batch Normalization, and ReLU activation, except for the last one. NM-Net is trained by Adam~\cite{adam2014} with a learning rate being $10^{-3}$ and batch size being 16. For LGC-Net, we use the code released by the authors to train the model. For PointNet and PointNet++, we adopt the ResNet backbone following the accomplishment in~\cite{yi2018learning}. To verify the effectiveness of compatibility-specific neighbor mining method, another version of NM-Net is also implemented (named NM-Net-sp), where compatibility-specific $knn$ search is replaced by spatially $knn$ search. Parameters including the number of neighbors $k$ and $\lambda$ (Eq.~\ref{eq:LRF6}) are set to 8 and $10^{-3}$, respectively.

\noindent\textbf{Benchmark datasets.}  Four datasets
are employed in our experiments - \ie, NARROW, WIDE, COLMAP~\cite{schoenberger2016sfm}, and MULTI~\cite{zhao2018scalable} (Table~\ref{tab:tab1}). The first two datasets are collected by us using a drone in four scenes, and we respectively keep a sample interval of 10 and 20 frames to attain narrow and wide baseline matching data. For NARROW, WIDE, and COLMAP, the ground-truth camera parameters are obtained by VisualSFM~\cite{Wu2013Towards} and the ground-truth labels of correspondences are calculated by comparing the corresponding epipolar distances~\cite{hartley2003multiple} with a threshold ($10^{-4}$). MULTI is a tiny dataset consisting of $45$ image pairs with available ground-truth labels for each correspondence. We use this dataset to test the generalization in cases of multi-consistency matching.

\noindent\textbf{Evaluation criteria.} To measure the correspondence selection performance, we employ precision (P), recall (R), and F-measure (F) as in~\cite{lin2014bilateral, bian2017gms, Ma2017Locality}. Moreover, considering that accurate estimation of the global transformation is required in some image alignment and 3D reconstruction tasks ~\cite{snavely2008modeling, benhimane2004real, brown2007automatic}, the deviation between the essential matrix $\mathbf{E}$ estimated by selected correspondences and ground-truth $\mathbf{E}_{gt}$ is measured by MSE, MAE, median, max, and min as in~\cite{ranftl2018deep}. Since P, R, and F explicitly reflect the performance of correspondence selection, we will focus on analyzing results from the perspective of these metrics.
\subsection{Single Consistency}
\label{sec:single}

Finding a single consistency corresponds to a global transformation (\eg, essential matrix) in static scenes is a popular application~\cite{snavely2008modeling, benhimane2004real} for feature correspondences. Our experimental results on NARROW, WIDE, and COLMAP datasets which contain a single consistency in each image pair are presented in Table~\ref{tab:tab2}.

\begin{figure}[t]
	\centering
	\subfigure[LGC-Net]
	{ \includegraphics[width=0.47\linewidth]{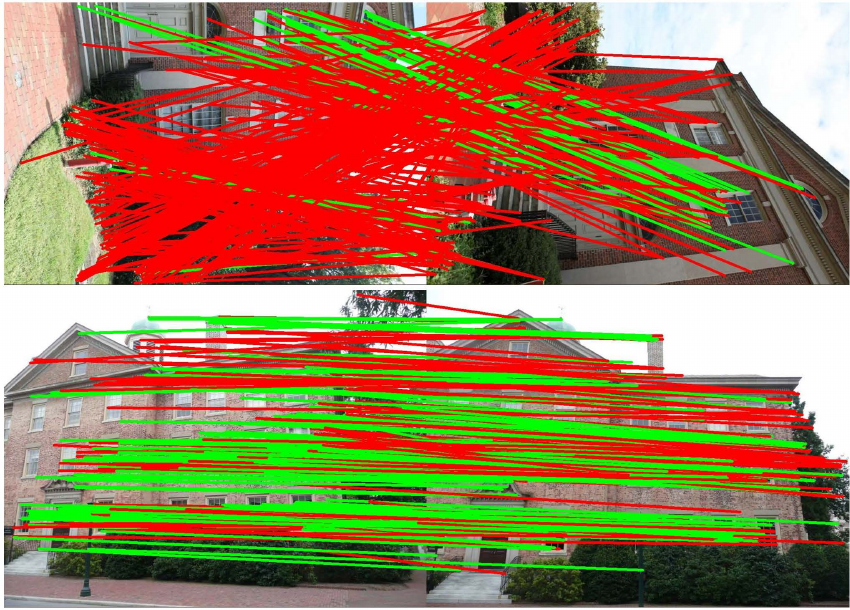}\label{fig_5a} }
	\subfigure[NM-Net]
	{ \includegraphics[width=0.47\linewidth]{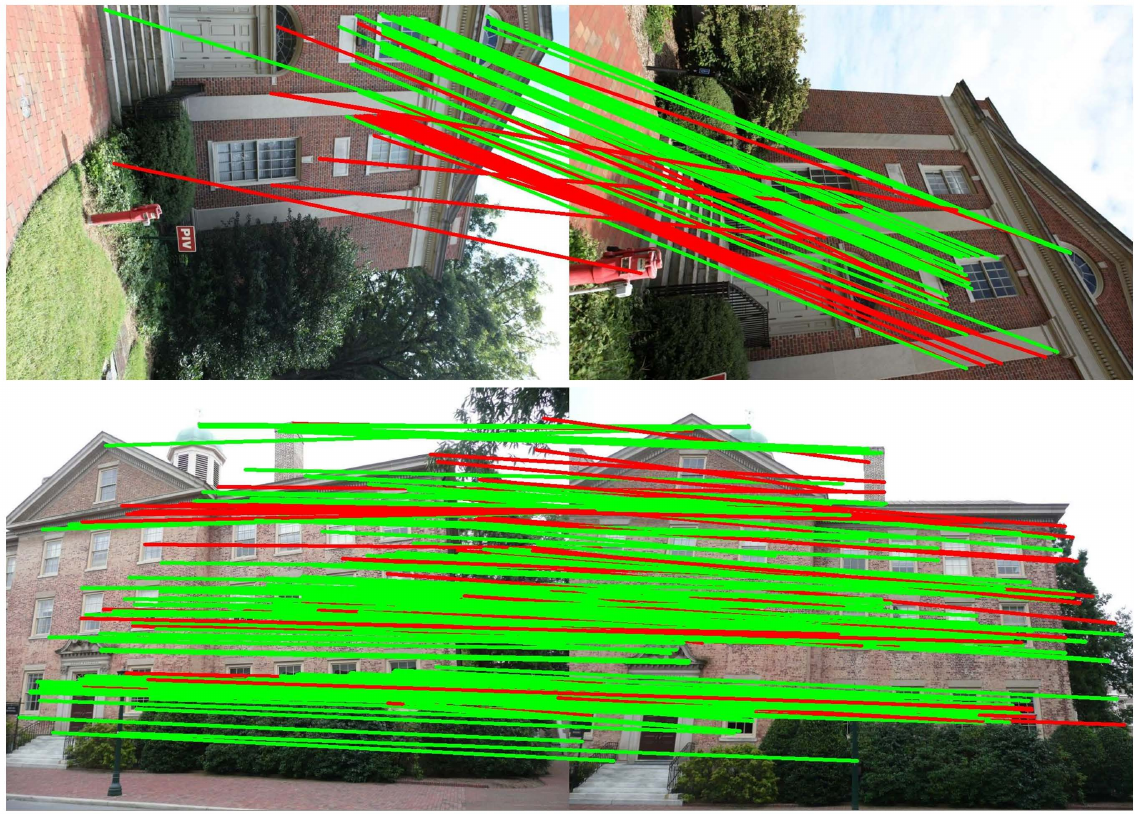}\label{fig_5b} }	
	\caption{{\bf Visual results on COLMAP dataset.} Green and red lines represent inliers and outliers in the selected correspondence set by LGC-Net and NM-Net, respectively.}
	\label{fig:5}
	\label{fig:onecol}
\end{figure}

As reported in Table~\ref{tab:tab2} (a), (b), and (c), NM-Net significantly outperforms hand-crafted algorithms and other learning-based approaches in terms of F-measure. First, when compared with hand-crafted algorithms such as RANSAC, GTM (using the same binary item in Eq.~\ref{eq:LRF6}), and LPM (employing spatially $k$-nearest information), NM-Net outperforms them by about 20 percentages on all three datasets. Second, NM-Net remarkably surpasses a representative set of learning-based approaches. For PointNet and LGC-Net, global features are extracted by average pooling and Context Normalization, respectively. For PointNet++, local information is added by spatially $knn$ search for each correspondence. NM-Net also extracts both global features and local features but mines neighbors relying on the proposed compatibility metric of Eq. \eqref{eq:LRF6}. The superiority of our framework can be easily verified. Third, NM-Net performs better than NM-Net-sp on all datasets; the gap becomes more dramatic on COLMAP which is a more challenging dataset with an extremely low initial inlier ratio ($7.496\%$). It implies that our compatibility-specific $knn$ search is more robust to high outlier ratios than standard spatially $knn$ search. Some representative visual comparison results are presented in Fig.~\ref{fig:5}, in which NM-Net is compared against current state-of-the-art deep learning framework LGC-Net. More visual results can be found in the supplementary material.

\subsection{Multiple Consistencies}
\label{sec:mul}

Multi-consistency feature matching in the situation of dynamic scenarios remains an open research problem as mentioned in~\cite{zhao2018scalable}. In contrast to a global transformation for static scenes, several local transformations corresponding to multiple consistencies are included into the initial correspondence set. Because MULTI only contains $45$ image pairs, models pretrained on NARROW that includes a similar inlier ratio (Table~\ref{tab:tab1}), are adopted to test the generalization from single consistency to multiple consistencies.

\begin{table}[t]
	\begin{center}
		\begin{tabular}{|c||c||c||c|}
			\hline
			Method & P (\%) & R (\%) & F (\%)\\
			\hline\hline
			PointNet~\cite{qi2017pointnet} & 48.223 & 5.829 & 8.717\\
			PointNet++~\cite{qi2017pointnet++} & {\bf64.661} & 7.871 & 13.327 \\
			LGC-Net~\cite{yi2018learning} & 61.736 & {\bf36.849} & {\bf41.849} \\
			NM-Net & 51.898 & 33.653 & 35.605\\
			\hline
		\end{tabular}
	\end{center}
	\caption{{\bf Generalization on MULTI dataset.} Tested models are pretrained on NARROW dataset. Metrics are precision (P), recall (R), and F-measure (F).}
	\label{tab:tab3}
\end{table}
\begin{figure}[t]
	\centering
	\subfigure[LGC-Net]
	{ \includegraphics[width=0.45\linewidth]{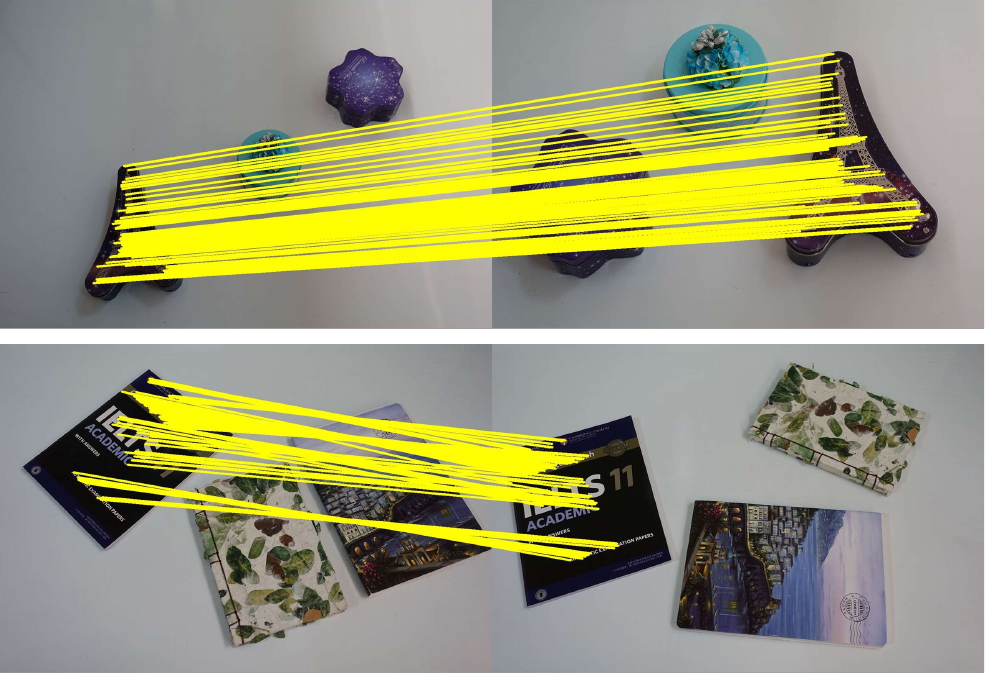}\label{fig_5a} }
	\subfigure[NM-Net]
	{ \includegraphics[width=0.45\linewidth]{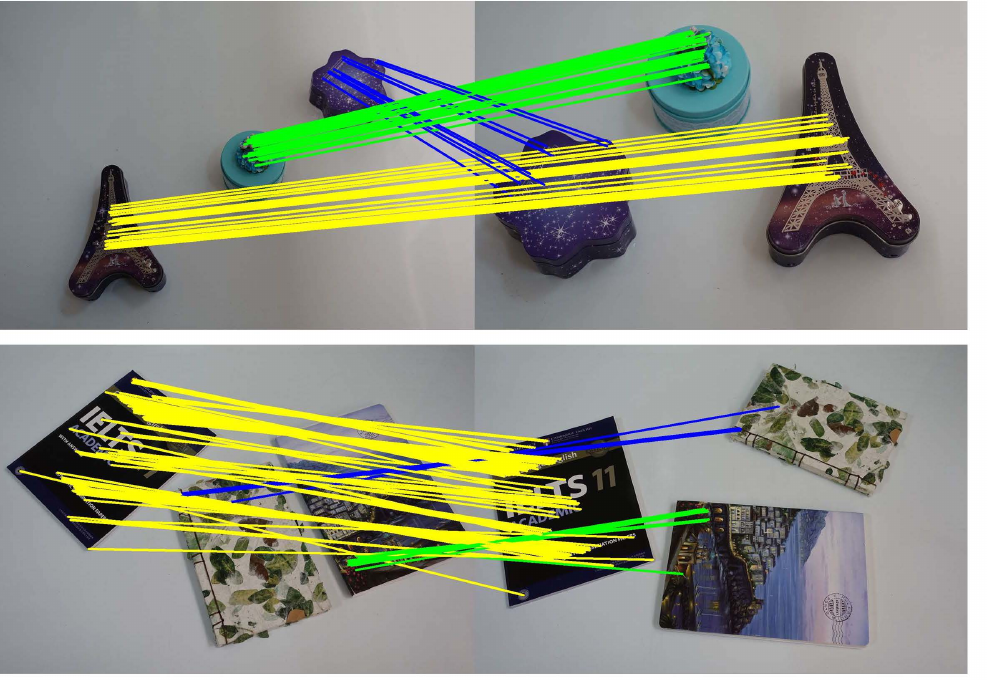}\label{fig_5b} }	
	\caption{{\bf Visual results on MULTI dataset.} Different colors represent different feature consistencies.}
	\label{fig:9}
	\label{fig:onecol}
\end{figure}
Table~\ref{tab:tab3} and Fig.~\ref{fig:9} show quantitative and qualitative results on MULTI dataset, respectively. Although LGC-Net achieves higher F-measure than NM-Net (still the second best), NM-Net is able to pick out all kinds of consistencies; while LGC-Net finds out only one kind of consistency. Note that LGC-Net is trained under the supervision of a hybrid loss that includes a regression loss corresponding to a global transformation, which explains why LGC-Net is less effective for multi-consistency matching with several local transformations. By contrast, NM-Net is trained with a  classification loss that is insensitive to multiple consistencies.
\subsection{Method Analysis}
\label{sec:ana}
\begin{figure*}[!ht]
	\centering
	\subfigure[NARROW]
	{ \includegraphics[width=0.3\linewidth]{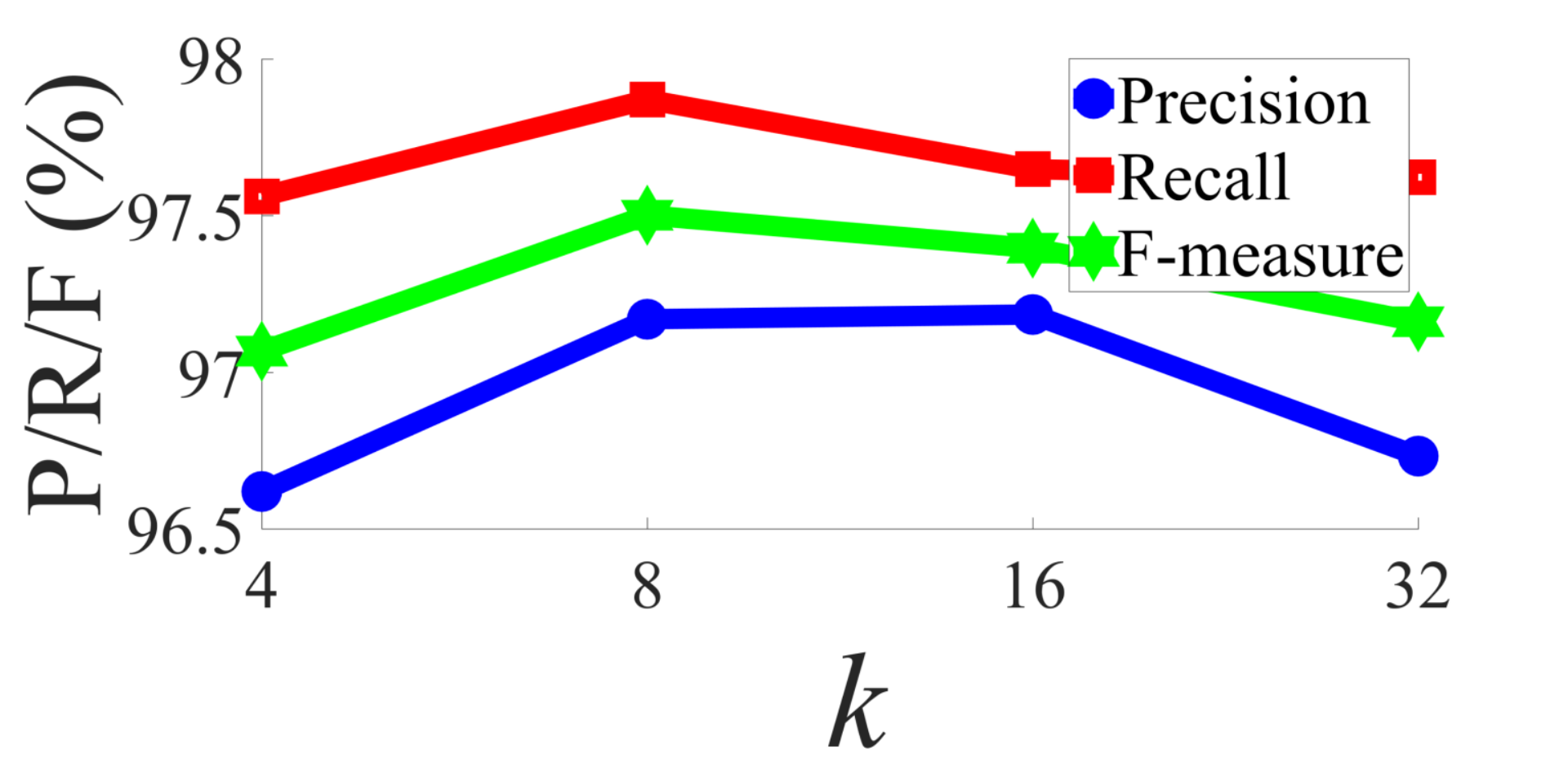}\label{fig_6a} }
	\subfigure[WIDE]
	{ \includegraphics[width=0.3\linewidth]{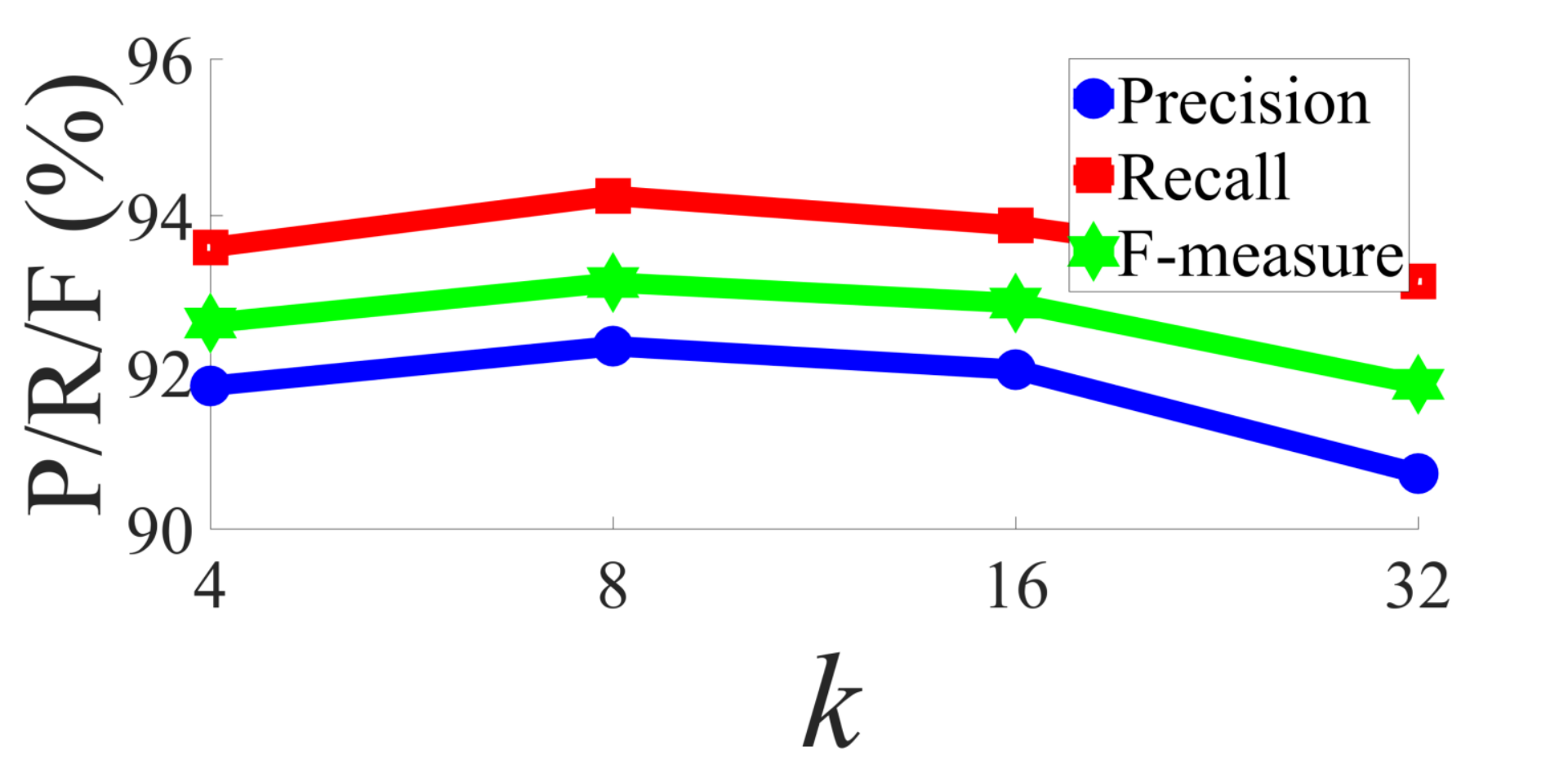}\label{fig_6b} }
	\subfigure[COLMAP]
	{ \includegraphics[width=0.3\linewidth]{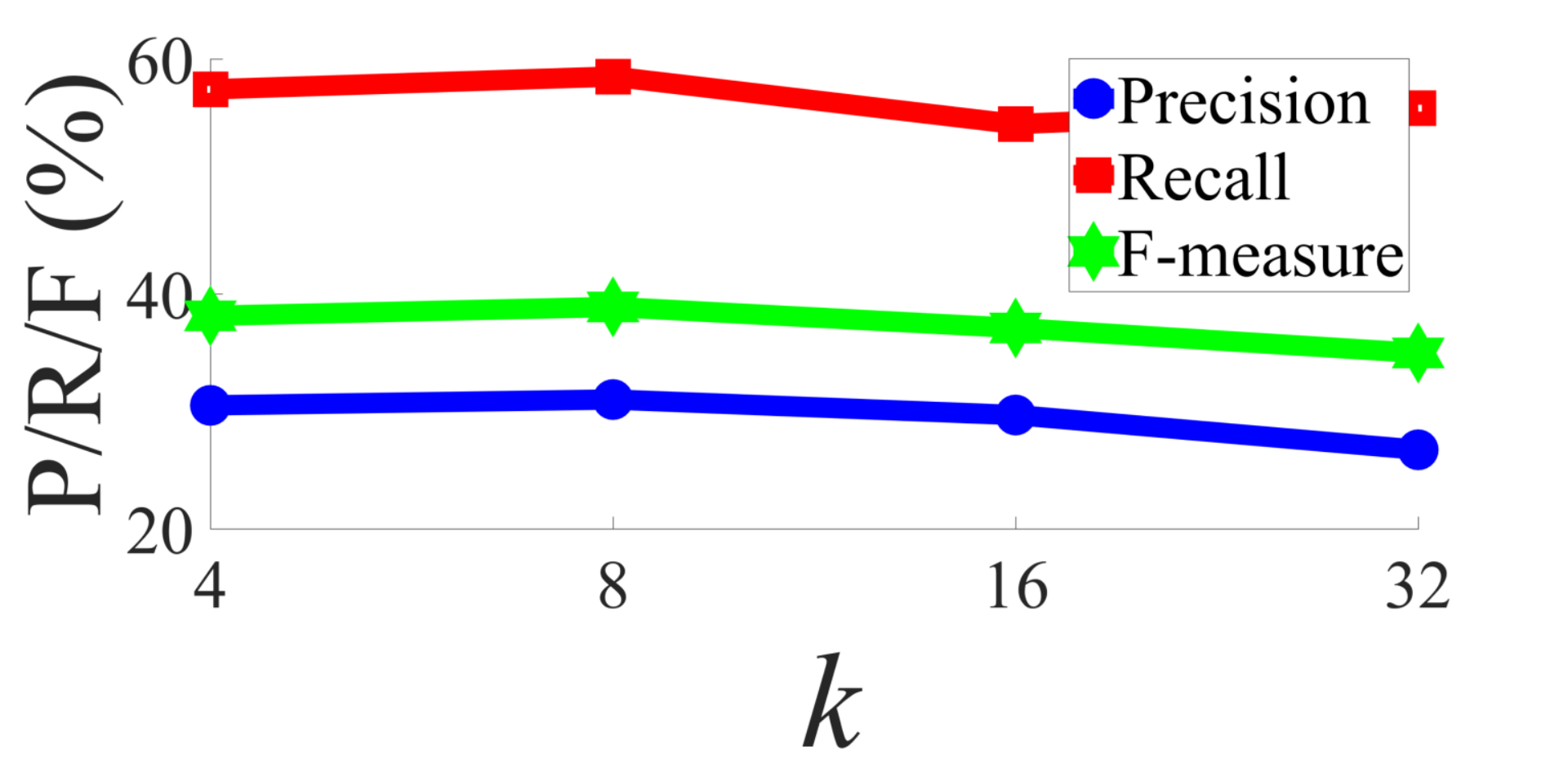}\label{fig_6c} }		
	\caption{{\bf Analysis of parameter $k$.} We train our NM-Net with different values of neighborhood size $k$ ($\left\lbrace 4, 8, 16, 32\right\rbrace$) while keep other settings identical, and examine the performance variation on NARROW, WIDE, and COLMAP datasets.}
	\label{fig:6}
	\label{fig:onecol}
\end{figure*}

\noindent\textbf{Parameter $k$.} The number of neighbors $k$ for each correspondence is a core parameter in NM-Net, which determines the receptive field of local features in each graph $\mathcal{G}$. Consequently, several versions of NM-Net with different numbers of neighbors are studied on NARROW, WIDE, and COLMAP. As shown in Fig.~\ref{fig:6}, NM-Net experiences performance degradation when $k$ is either too small (4) or too large (32). When $k=4$, the search space for each correspondence is too small to extract enough available local features; when $k=32$, the consistency of neighbors in $\mathcal{G}_{inlier}$ for an inlier $c_{inlier}$ will decline and some outliers will be included undesirably as nuisance. Based on above analysis, we have used $k=8$ as the default size of neighbors in NM-Net.

\begin{figure}[t]
	\begin{center}
		\includegraphics[width=0.6\linewidth]{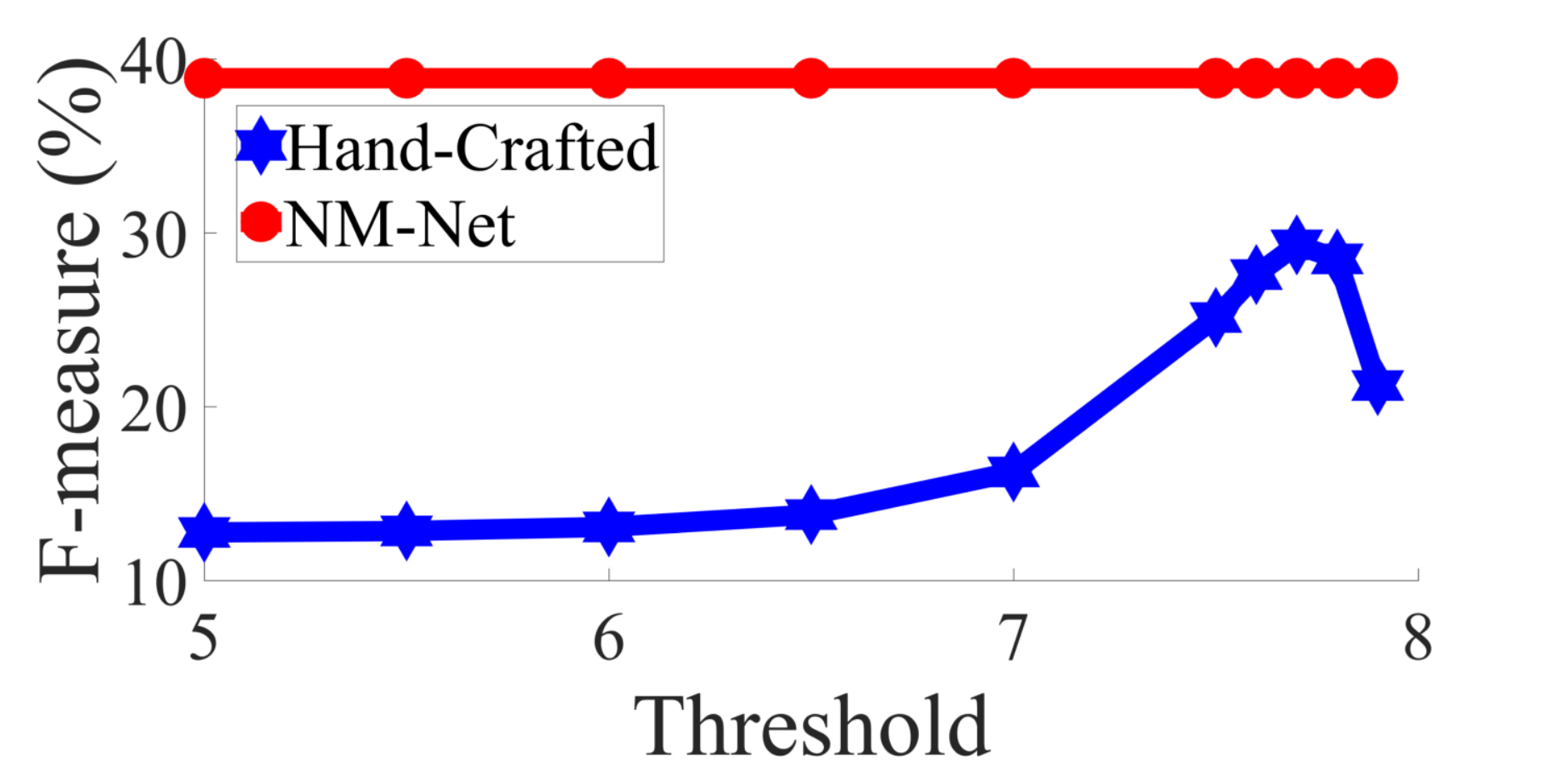}	
	\end{center}
	\caption{{\bf Learning effectiveness validation.} The hand-crafted method judges the correctness of a correspondence by comparing the sum of compatibility scores of the neighbors with a threshold. This threshold is varied from 5 to 7.9 to alleviate the effect of threshold setting.}
	\label{fig:7}
	\label{fig:onecol}
\end{figure}

\noindent\textbf{Learning effectiveness validation.} Since the neighbors searched by the compatibility metric are more consistent for inliers than for outliers, compatibility scores of correspondences in $\mathcal{G}_{inlier}$ should be higher than those in $\mathcal{G}_{outlier}$ in theory. A potential issue arises: can correspondences be directly classified based on the scores? To address this question, a hand-crafted approach is designed to calculate the sum of scores in each graph $\mathcal{G}_i$ as $s_{i}^{sum}$. Then $c_i$ will be determined as an inlier if $s_{i}^{sum}$ is higher than a predefined threshold. A comparison between the hand-crafted approach and NM-Net on COLMAP dataset is included in Fig.~\ref{fig:7}. Clearly, NM-Net achieves a far higher F-measure than the hand-crafted one with all thresholds. Due to uncertainties such as viewpoint changes and camera rotation, the distribution of correspondences is distinctly different among various image pairs. Utilizing raw scores to distinguish inliers and outliers is therefore unreliable. In contrast to the hand-crafted approach, the scores are leveraged as indexes to search for compatible neighbors in our proposed method. The local features hidden in these neighbors can be fully explored by a powerful deep learning network.

\noindent\textbf{Compatibility metric analysis.} Since compatibility-specific $knn$ search is a key procedure in NM-Net to generate the initial graph $\mathcal{G}$, the consistency of elements in $\mathcal{G}$ directly affects the effectiveness of raw information. To shed more light to this important matter, an analysis of the compatibility metric (Eq.~\ref{eq:LRF6}) is conducted as follows.

\begin{figure}[t]
	\subfigure[Neighbors of inliers]
	{ \includegraphics[width=0.47\linewidth]{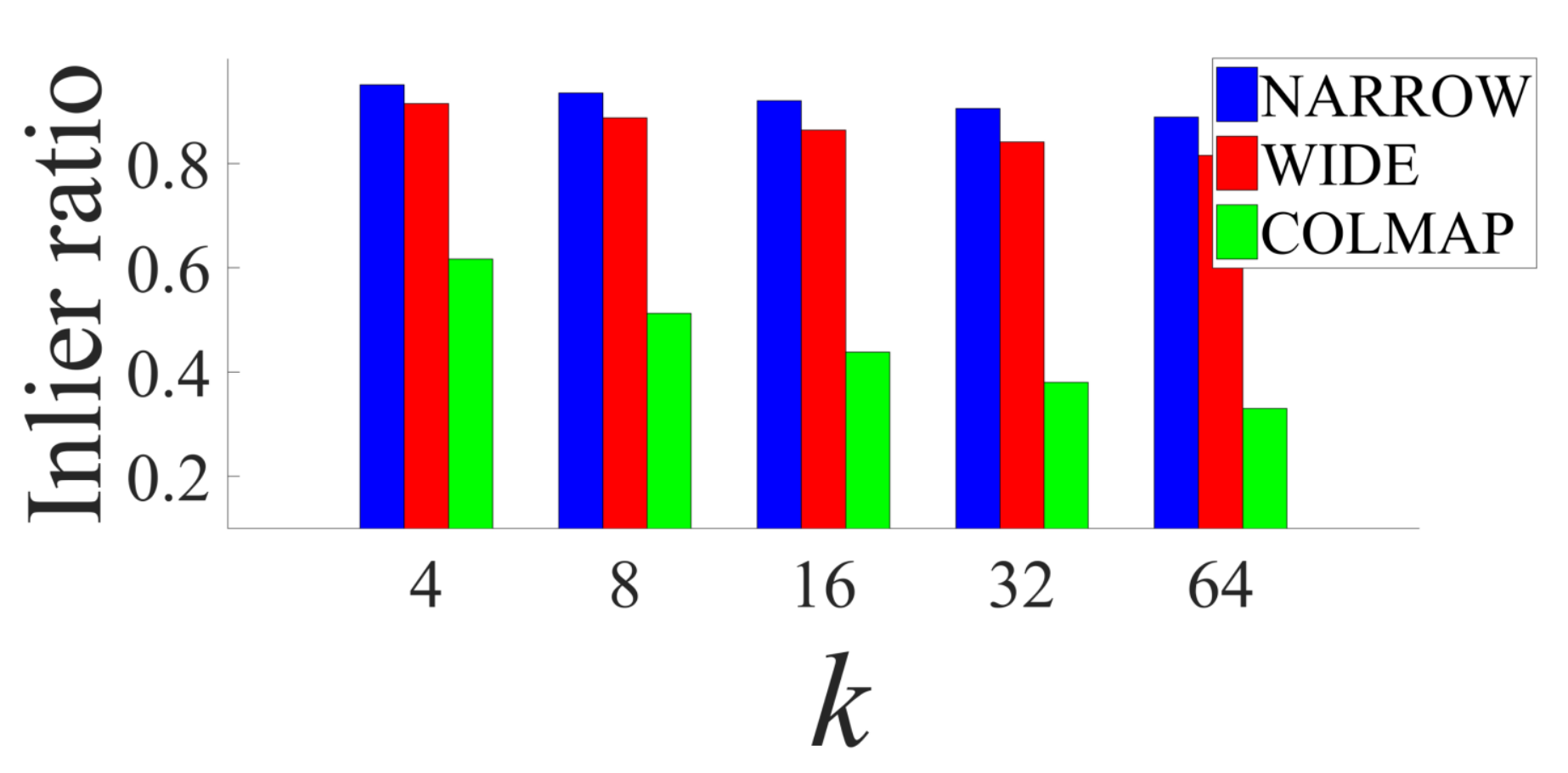}\label{fig_7a} }	
	\subfigure[Neighbors of outliers]
	{ \includegraphics[width=0.47\linewidth]{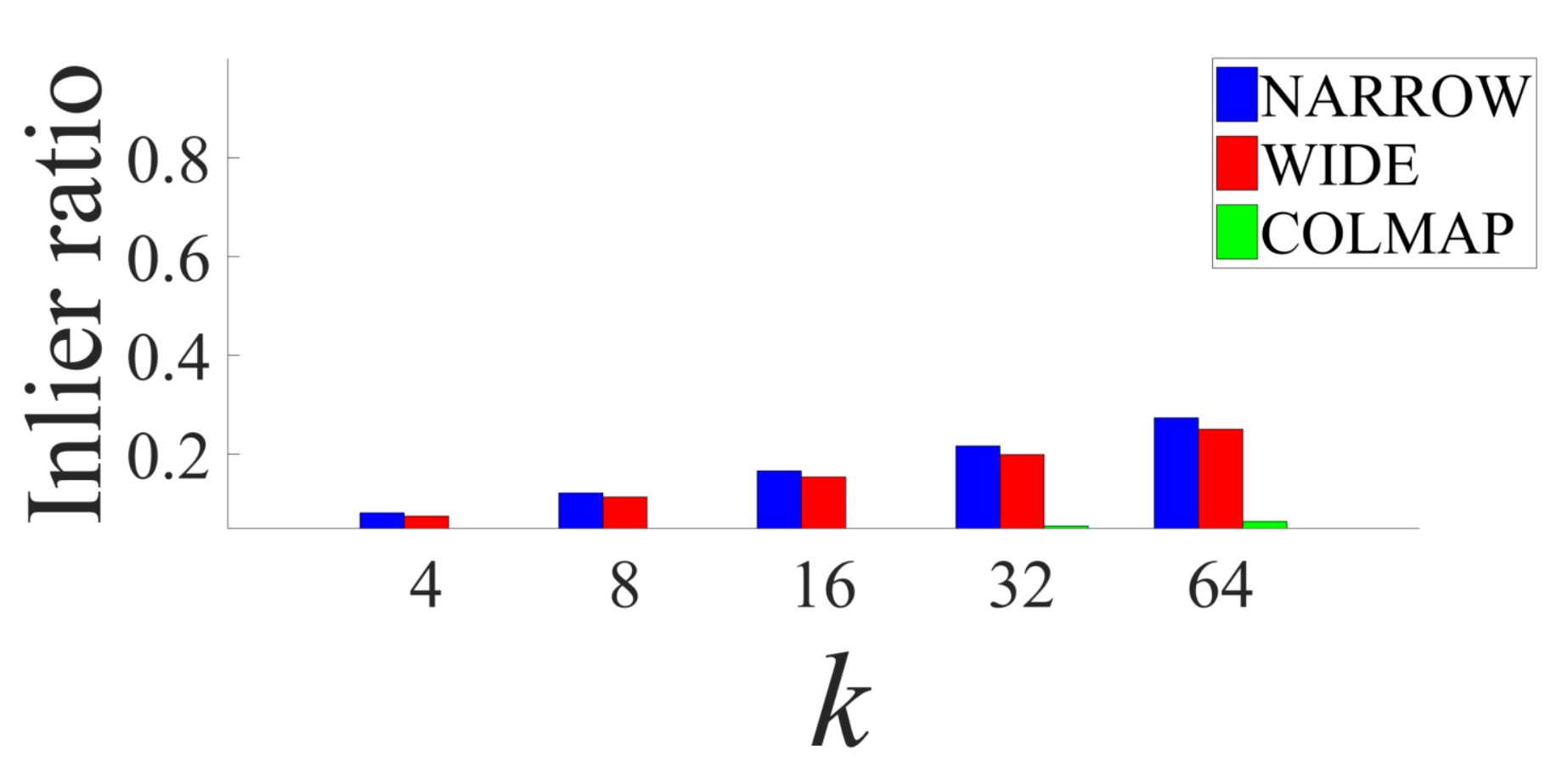}\label{fig_7b} }
	\caption{{\bf Compatibility metric analysis.} The inlier ratios of neighbors recognized via our compatibility metric (Eq.~\ref{eq:LRF6}) of (a) inliers and (b) outliers are calculated on NARROW, WIDE, and COLMAP datasets to examine if this metric can provide distinguishable local information.}
	\label{fig:8}
	\label{fig:onecol}
\end{figure}

The inlier ratios of the neighbors searched by Eq.~\ref{eq:LRF6} on NARROW, WIDE, and COLMAP datasets are shown in Fig.~\ref{fig:8}. The compatibility metric is deemed to be reasonable for the following reasons. First, the neighbors of inliers are significantly more consistent than the ones of outliers, with the inlier ratios being $3$ times higher. Second, our approach achieves approximate $100\%$ inlier ratios on both NARROW and WIDE datasets as shown in Fig.~\ref{fig:8} (a), where almost all searched neighbors of inliers are consistent (\ie, inliers). However, the inlier ratios of the neighbors of correct matches drop considerably on the more challenging COLMAP dataset. This result suggests that there is still a large room for further improvement from the perspective of robustness of the compatibility metric. We leave this to our future study.
\section{Conclusion}
We have presented a hierarchical classification network named NM-Net to select correct matches from initial correspondences, which fully mines compatibility-specific locality for each correspondence. Experiments demonstrate that NM-Net behaves favorably to the state-of-the-art (both hand-crafted and learning-based) approaches. The current shortcoming of our approach is the demand for key-point detectors with local affine information to compute the compatibility score. We expect developing a more advanced compatibility metric without such constraint in our future works.

\noindent\textbf{Acknowledgments. } This work was supported in part by the National Natural Science Foundation of China under Grant 61876211 and by the 111 Project on Computational Intelligence and Intelligent Control under Grant B18024.

{\small
\bibliographystyle{ieee}
\bibliography{bibfile}
}

\end{document}